\documentclass{article} %
\usepackage{iclr2024_conference,times}
\usepackage{inconsolata}

\usepackage{amsmath,amsfonts,bm}

\def\eqref#1{equation~\ref{#1}}

\def\1{\bm{1}}

\DeclareMathAlphabet{\mathsfit}{\encodingdefault}{\sfdefault}{m}{sl}
\SetMathAlphabet{\mathsfit}{bold}{\encodingdefault}{\sfdefault}{bx}{n}

\DeclareMathOperator*{\argmax}{arg\,max}

\usepackage{hyperref}
\usepackage{booktabs}
\usepackage{url}
\usepackage[capitalize]{cleveref}
\usepackage{graphicx}
\usepackage{xcolor}
\usepackage{mathtools}
\usepackage{wrapfig}
\usepackage{tikz}

\usepackage{xspace}
\usepackage{amsthm}
\usepackage[small]{caption}
\usepackage{subcaption}

\newcommand{\colorPatch}[2][xxxx]{
  \colorbox[HTML]{#2}{{\color[HTML]{#2}#1}}}
  
\newcommand{\colorContext}[4]{
  \framebox{\negthickspace\colorPatch{#1}} & \colorPatch{#2} & \colorPatch{#3} & \emph{#4}}

\newcommand{\xrollout}{\tau}
\newcommand{\xagent}{\pi^*}
\newcommand{\xmagent}{\pi}
\newcommand{\xinference}{\xmagent}
\newcommand{\xourmodel}{\xmagent}

\newcommand{\xstate}{s}
\newcommand{\xaction}{a}
\newcommand{\xreward}{R^*}
\newcommand{\xmreward}{R}

\newcommand{\xbudget}{\beta}

\newcommand{\xtemp}{\xbudget_{\text{temp}}}
\newcommand{\xdepthbudget}{\xbudget_{\text{runtime}}}

\newcommand{\budgetdist}{p_{\text{budget}}}
\newcommand{\xc}{\xbudget_{\text{puct}}}
\newcommand{\algo}{{latent inference budget model}\xspace}

\newcommand{\acronym}{L-IBM\xspace}
\newtheorem{proposition}{Proposition}
\newtheorem{definition}{Definition}

\title{Modeling Boundedly Rational Agents with \\ Latent Inference Budgets}

\author{Athul Paul Jacob \\
        MIT \\
\texttt{apjacob@mit.edu} \\
\And
Abhishek Gupta \\
University of Washington\\
\texttt{abhgupta@cs.washington.edu} \\
\And
Jacob Andreas \\
MIT \\
\texttt{jda@mit.edu}
}

\iclrfinalcopy %
\begin{document}

\maketitle

\begin{abstract}
We study the problem of modeling a population of agents pursuing unknown goals subject to unknown computational constraints. In standard models of bounded rationality, sub-optimal decision-making is simulated by adding homoscedastic noise to optimal decisions rather than explicitly simulating constrained inference. In this work, we introduce a \emph{\algo} \emph{(\acronym)} that models agents' computational constraints explicitly, via a latent variable (inferred jointly with a model of agents' goals) that controls the runtime of an iterative inference algorithm. \acronym{}s make it possible to learn agent models using data from diverse populations of suboptimal actors. In three modeling tasks---inferring navigation goals from routes, inferring communicative intents from human utterances, and predicting next moves in human chess games---we show that \acronym{}s match or outperform Boltzmann models of decision-making under uncertainty. Inferred inference budgets are themselves meaningful, efficient to compute, and correlated with measures of player skill, partner skill and task difficulty. 
\end{abstract}

\section{Introduction}
Building effective models for multi-agent decision-making---whether cooperative or adversarial---requires understanding other agents' goals and plans. To help a friend navigate in a new environment, we must first understand where they want to go; to beat an opponent at chess, we must be able to predict their likely next moves. 
But decision-making, in humans and machines, is subject to computational constraints. Decision-makers often act suboptimally, relying on heuristics and approximations to choose their actions. Techniques that do not account for this suboptimality carefully may attribute behavior to differing intentions rather than different inference procedures.

How should we interact with agents seeking to accomplish unknown goals subject to unknown computational constraints? 
In standard models of bounded rationality \citep{luce2012individual}, sub-optimal decision-making is simulated by adding noise to optimal decisions rather than explicitly simulating constrained inference. This results in models that treat agents as uniformly suboptimal in a way that fails to account for sub-optimal inference \emph{algorithms} or for \emph{non-homogenous} suboptimality.

In this paper, we describe a simple approach for building models of agents given traces of their behavior. 
Our approach explicitly models agents' ``inference budgets'',
via a latent variable that controls the runtime of each agent's inference procedure.
We show that for agents performing inference using \emph{anytime algorithms} (algorithms that can be terminated at any point and return approximately correct solutions) inference budgets can be efficiently inferred from example behaviors.
A diverse set of multi-agent decision-making procedures---including graph-based planning algorithms, recursive-rational models of human language production, and Monte Carlo tree search---admit imputation of inference budgets in this framework.

In three diverse agent modeling tasks---inferring navigation goals from routes, inferring communicative intents from human utterances, and predicting subsequent moves in human--human chess matches---we show that our approach matches or outperforms Boltzmann models of decision-making under uncertainty. Moreover, inferred inference budgets are themselves meaningful, correlating with measures of player skill, partner skill, and task difficulty. Our results show that sub-optimal human decision-making can be efficiently modeled with computationally constrained versions of standard search algorithms. By doing so, we obtain both accurate models of humans' decision-making and informative measures of their inferential capacity.
\section{Background and problem formulation}
\label{sec:background}
\begin{wrapfigure}{r}{0.5\textwidth}
\vspace{-1em}
\includegraphics[width=0.5\textwidth,trim=.2in 4.5in 4.3in .2in]{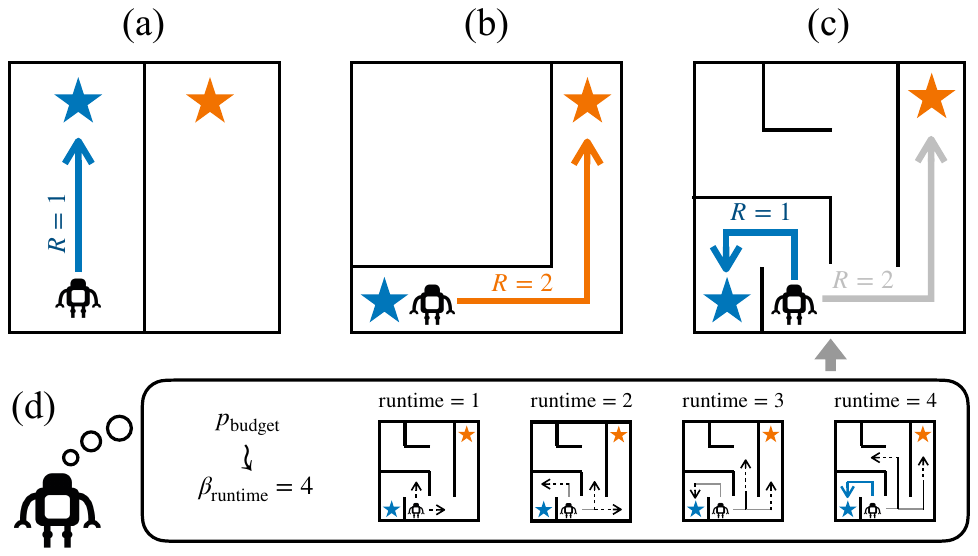}
\caption{Inferring rewards from boundedly-rational trajectories. The agent will move to the blue star (a), but prefers to move toward the orange star when both are available (b). When locating the orange star requires solving a harder search problem, however, the agent seeks the blue star instead, indicating that its search abilities are limited (c). 
Our proposed approach automatically infers the budget that the agent uses when planning (d). 
Knowing this budget, we could perhaps assist this agent by providing a targeted hint (\emph{move right}) at the beginning of its trajectory.}
\vspace{-1.5em}
\label{fig:toy-example}
\end{wrapfigure}

We study the problem of modeling one or more agents given given traces of their behavior. 
In particular, we assume that we observe a collection of trajectories (state--action sequences) produced by agents $\xagent : \xstate \mapsto \xaction $ acting in a Markov decision process to maximize some reward function $\xreward(\tau)$.
Even when $\xreward(\tau)$ is known to agents, inferring optimal actions is often intractable, so agents in the real world will in general \emph{approximate} optimal behavior subject to some (unknown) computational constraints (which may differ from agent to agent). From this data, we seek to infer \textbf{agent models} $\xmagent$ defined in terms of (1) estimates $\xmreward$ of reward function $\xreward$ (known to agents but not modelers), and (2) descriptions of the computational limitations that govern agents' choice of actions. In other words, we seek to model both \emph{what agents wish to do} and \emph{what agents will actually do} in any given state. 
\cref{fig:toy-example} shows a conceptual example: assuming the agent receives a different reward for reaching each of the two goals,
the three trajectories depicted there cannot be generated by the optimal policy for any reward function, but can be explained by model that can only look ahead to a limited number of positions in the maze.

Throughout this paper, we will model agent actions as arising from an \textbf{approximate inference procedure} $\xinference(a \mid s; \xmreward, \xbudget)$ that takes as input a reward function and a \textbf{computational budget} $\xbudget$. We model agents by inferring values of $\xmreward$ and $\xbudget$ given the executed trajectories $\tau_i$.

The ability to infer goals from suboptimal (and even completely unsuccessful) plans is a key human skill, present in children as young as 18 months \citep{meltzoff1995understanding}. Computational models of bounded rationality thus have a long history in artificial intelligence, cognitive science, and behavioral economics. But what does this suboptimality look like in practice, and how should we model and infer the inference budget $\xbudget$ simply from observations of behavior?

One of the most widely used models of boundedly rational decision-making is the so-called \textbf{Boltzmann} model \citep{luce2012individual}, in which agents take actions according to
\begin{equation}
\label{eq:boltzmann}
    \xinference(\xaction \mid \xstate; \xmreward, \xbudget) \propto \exp \{ \beta \cdot \xmreward(\xstate, \xaction) \}
\end{equation}
This equation has a number of appealing interpretations, e.g.\ as the policy that achieves a target reward while maximizing entropy. It has been used to model not just the selection of actions, but also trajectories, preferences, corrections, and more---see \citep{jeon2020reward} for a recent survey. More elaborate approaches in this family also predict $\xbudget$ conditioned on the current state or action history, making it possible to model state-dependent skill \citep{beliaev2022imitation}.

However, Boltzmann models have a significant limitation: the probability of generating an action in \cref{eq:boltzmann} depends only on the true value of that action, and not on the cost of acquiring a high-quality value estimate in the first place. To see why this might be a problem, consider again the trajectories depicted in the conceptual example \cref{fig:toy-example}(b--c), which differ \emph{only} in the difficulty of the search problem, and not in the cost of the optimal trajectory at all. A model of boundedly rational decision-making with the form of \cref{eq:boltzmann} cannot account for this difference.

There is a large body of other approaches on modeling human planning under resource constraints in psychology, economics and in classical AI \citep[][inter alia]{
callaway2022rational, 
russell1991principles,
huys2015interplay, 
huys2012bonsai, 
camerer2004cognitive, 
griffiths2019doing, 
boddy1989solving,
nature2023expertise}. 
However, these approaches make strong assumptions about how planning is performed, limiting their applicability to real-world data. Here, we seek to develop a general framework that avoids strong assumptions about either the functional form of the reward model or the algorithmic form of the planning procedure.
As a result, we can apply this single framework to real-world behavior in tasks as diverse as language generation and chess gameplay.

\section{Inferring Rewards and Inference Budgets from Behavior}
\label{sec:method}

As motivated in \cref{sec:background}, our goal is to model agents acting to optimize an unknown value function subject to an unknown computational constraint. In practice, we often want to model populations comprising multiple agents or agent sub-populations $(\xagent_1, \xagent_2, \ldots \xagent_N)$ with a shared reward function $\xreward$ (e.g.\ winning at chess) but differing computational constraints. 

To do so, we assume we have access to a collection of trajectories $\{\xrollout\}_i = \{\xrollout^1_i, \xrollout^2_i, \ldots \xrollout^{M_{i}}_i\}$, with each collection of trajectories $\{\xrollout\}_i$ generated by a different agent or sub-population $i$. 
We model these trajectories as drawn from the following generative process: 
\begin{enumerate}
    \item at each timestep, agent $i$ draws a budget $\xbudget$ from an agent-specific prior $\budgetdist(\xbudget\mid\eta_i)$
    \item $\xagent_i$ chooses actions according to a budget-constrained inference procedure $\xagent_i(a \mid s; \xreward, \xbudget)$ 
\end{enumerate}
Because budgets may vary between trajectories,
learning a model of these agents ultimately consists of learning reward parameters $\theta$ and agent-specific budget-generating parameters $\eta_i$ while \emph{marginalizing} over latent budgets themselves. We do so via maximum \emph{a posteriori} inference, optimizing:

\begin{align}
\label{eq:objective}
\argmax_{\theta, \eta} \sum_{\substack{
i \\
\tau \in \{\tau\}_i \\
(s, a) \in \tau
}}
    \log \xmagent(\xaction \mid \xstate; \theta, \eta) = \argmax_{\theta, \eta} 
    \sum_{\substack{
i \\
\tau \in \{\tau\}_i \\
(s, a) \in \tau
}}
    \log \sum_{\xbudget} \budgetdist(\xbudget\mid\eta_i) \cdot \xinference(a \mid s; \xmreward_\theta, \xbudget)
\end{align}

If $\xinference(a \mid s; \xreward, \xbudget)$ is an arbitrary inference algorithm, \cref{eq:objective} might present a challenge: this inference procedure must be run for all possible values of $\xbudget$, which will in general be intractable. Under what circumstances can we optimize this equation efficiently? The key observation in this paper is that if $\xinference$ is an \emph{anytime inference algorithm} \citep{anytime1988}, we can evaluate $n$ values of $\xbudget$ as quickly as we can evaluate one, making this optimization tractable. 
\begin{definition}
An anytime algorithm $\xinference$ is one that runs for $t$ timesteps and produces a sequence of inference states $(f_1, f_2, \ldots f_t)$, where every $f_{i}$ can be computed from $f_{i-1}$ in $\mathcal{O}(1)$ time, and $f_i$ can be used to select an action according to some $\xinference(a \mid s; \xmreward, f_i)$.
\end{definition}
As we will see shortly, many canonical inference algorithms used in single- and multi-agent decision-making scenarios have this from.
In these cases, rather than letting the budget parameter $\xbudget$ determine noise or suboptimality, we may use it to parameterize the runtime of the agent's inference procedure itself, writing:
\begin{equation}
    \log \xmagent(\xaction \mid \xstate; \theta, \eta_i) = \log \sum_\xbudget \budgetdist(\xdepthbudget \mid \eta_i) \cdot \xmagent(\xaction \mid \xstate; \xmreward_\theta, f_{\xdepthbudget})
\end{equation}
where we have denoted the budget $\xdepthbudget$ to indicate that it parameterizes the runtime of the anytime inference algorithm.
Crucially---by definition---computing this sum up to some maximum $\xbudget$ requires no more time than computing its final term.

The remainder of this paper looks at instantiations of this basic modeling framework in three different domains. In \cref{sec:maze}, we study the problem of inferring navigation goals from maze domain using a truncated graph search algorithm. In \cref{sec:prag}, we study rational speech acts (RSA) for inferring communicative intents from human utterances. Finally, in \cref{sec:chess}, we model human action prediction in chess using Monte-Carlo tree search (MCTS).

\vspace{-0.5em}
\section{Solving Mazes with Truncated Depth-First Search}
\vspace{-0.5em}
\label{sec:maze}

\begin{figure}[h]
\begin{minipage}{0.16\textwidth}
  \centering
  \includegraphics[width=\linewidth]{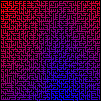}
  \subcaption{}
  \end{minipage}
  \hfill
  \begin{minipage}{.16\textwidth}
    \centering
    \includegraphics[width=\linewidth]{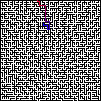}
      \subcaption{}
  \end{minipage}%
  \begin{minipage}{.16\textwidth}
    \centering
    \includegraphics[width=\linewidth]{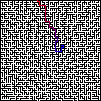}
    \subcaption{}
  \end{minipage}%
  \begin{minipage}{.16\textwidth}
    \centering
    \includegraphics[width=\linewidth]{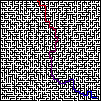}
    \subcaption{}
  \end{minipage}%
  \begin{minipage}{.16\textwidth}
    \centering
    \includegraphics[width=\linewidth]{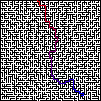}
    \subcaption{}
  \end{minipage}%
    \begin{minipage}{.16\textwidth}
    \centering
    \includegraphics[width=\linewidth]{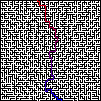}
    \subcaption{}
  \end{minipage}%
  \caption{Examples of the maze task. (a) Example of the value function heuristic applied to each state in the maze. Red indicates low value states and blue indicates high value states. (b)-(f) depicts the example trajectories of agents with depth budgets of 1, 2, 5, 10 and 20.}
  \label{fig:solved_mazes}
\end{figure}

We begin with a pedagogical example of \algo applied to a simple, single-agent decision-making task: maze navigation. 
Agents are placed at a random position in a maze with five exits. Each exit is a state $e_i$ associated with a reward $R_i$. Agents attempt to navigate toward the highest scoring exit by taking navigation actions (\texttt{north}, \texttt{east}, \texttt{south}, \texttt{west}). Here our goal is to recover the rewards $R_i$ that a single agent associates with each exit, along with agent budget parameters $\eta$, given observations of the agent's behavior.

\subsection{Agent Model}
\label{sec:maze-agents}

We assume that agents select navigation actions using a heuristic with a known functional form, in which the value of a state $\xstate$ is approximated as:
\begin{equation}
    V(\xstate) = \frac{
    \sum_i  R_i e^{-\|\xstate - e_i\|_1 \cdot R_i}
    }{
    \sum_i e^{-\|\xstate - e_i\|_1 \cdot  R_i}
    }
\end{equation}
where $\| s - s' \|_1$ measures the Manhattan distance between a pair of states (i.e.\ maze positions).
Intuitively, we model agents as ``attending'' to each exit in proportion to both its distance and associated reward. %
We assume that agents use this heuristic to perform \textbf{truncated breadth-first search}. In a state $s$, agents first estimate the value of each action $a$ by computing the value of the best state reachable in $\xdepthbudget$ actions, starting with $a$. Formally:
\begin{equation}
    Q_{\text{runtime}}(a \mid s) = \max_{\tau : \tau_0 = a, |\tau| = \xdepthbudget} V(\tau_{\xdepthbudget})
\end{equation}
where $\tau_0$ and $\tau_{\xdepthbudget}$ respectively denote the first and last actions in the trajectory $\tau$. Finally, agents select actions in proportion to these Q-values \citep{haarnoja2017reinforcement}: 
\begin{equation}
\label{eq:nts_policy}
\pi(a \mid s; \xdepthbudget, R) \propto e^{Q(a \mid s)}
\end{equation}
With this agent parameterization, \cref{eq:objective} can be computed efficiently:
\begin{proposition}
Truncated breadth-first Search (TBFS) is an anytime inference algorithm. (Represent each inference state $f_\beta$ as the set of frontier states and values reachable from each starting action. To compute $f_{\beta+1}$, add the unexplored children of these states to the set.)
\end{proposition}
\subsection{Data}

In this pedagogical example, we treat the agent model in \cref{sec:maze-agents} as the true data-generating process. We fix a set of parameters $R_i$ and $\xdepthbudget$, generate a collection of synthetic trajectories using \cref{eq:nts_policy}, then attempt to recover these parameters using \cref{eq:objective}. (This allows us to validate the feasibility of our approach under ideal conditions---later sections will apply it to real datasets of human-generated actions). In particluar, we generate 5 agents with runtime budgets of 1, 2, 5, 10, and 20 respectively. Example trajectories from each of these agents are depicted in \cref{fig:solved_mazes}.

\subsection{Evaluation}

\begin{figure*}[t]
    \begin{minipage}[b]{0.7\textwidth}
        \centering
        \begin{minipage}{0.45\textwidth}
        \centering \small \textbf{Inferred} $\xtemp$
        \end{minipage}%
        \vspace{1em}
        \begin{minipage}{0.45\textwidth}
        \centering \small \textbf{Inferred} $\xdepthbudget$ \\
         \centering (L-IBM)
        \end{minipage}
        \begin{minipage}{0.45\textwidth}
            \centering
            \includegraphics[width=\textwidth]{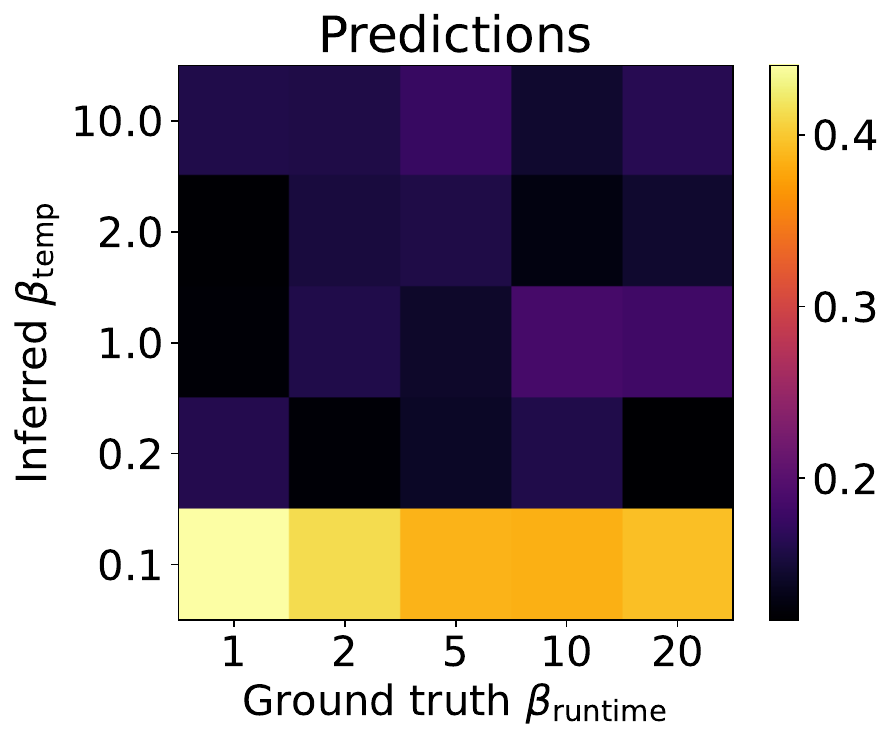}
            \subcaption{}
            \label{fig:maze_temp}
        \end{minipage}%
        \begin{minipage}{0.45\textwidth}
            \centering
            \includegraphics[width=\textwidth]{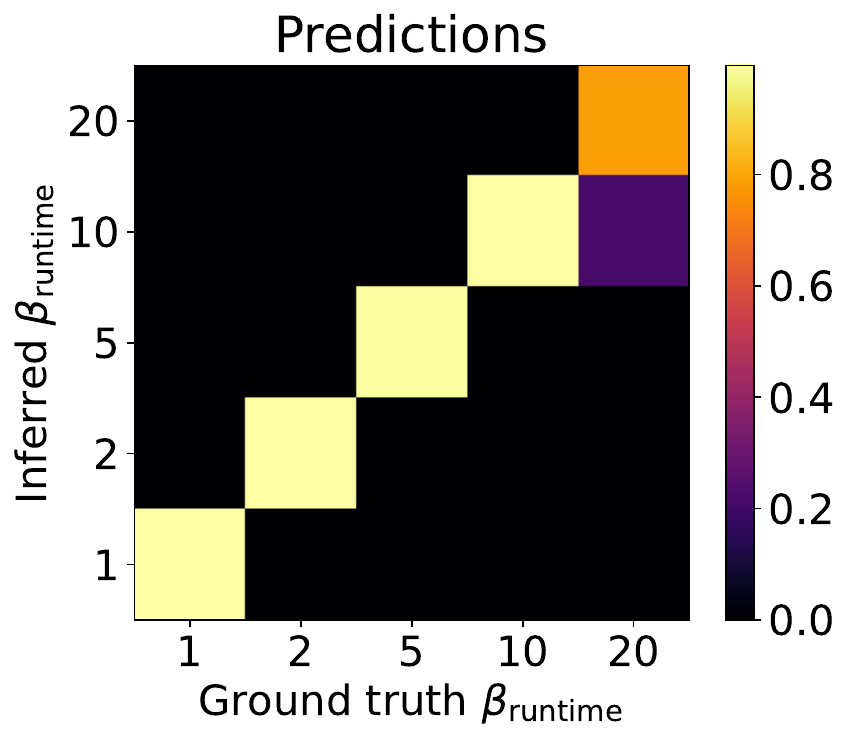}
            \subcaption{}
            \label{fig:maze_depth}
        \end{minipage}
        \vspace{-0.5em}
    \caption{Inferred parameters $\eta_i$ (distributions over $\xbudget$) for the maze navigation task. (a) \acronym almost perfectly recovers these parameters, while (b) the Boltzmann model shows no significant differences across inferred $\xtemp$.}
    \end{minipage}%
    \hspace{0.01\textwidth} %
    \begin{minipage}[b]{0.25\textwidth}
                
    \begin{minipage}[b]{1.0\textwidth}
    \footnotesize
    \centering
            \resizebox{\linewidth}{!}{
            \setlength{\tabcolsep}{1pt}
    \begin{tabular}{lc}
            \toprule
            \textbf{Approach} & \textbf{Accuracy} \\
            \midrule
            $\xdepthbudget=0$ & 5 \\
            $\xdepthbudget=20$ & 16 \\
            \hline
            Inferred $\xtemp$ & 20 \\
            \acronym & \textbf{44} \\
            \bottomrule
        \end{tabular}}
    \captionof{table}{\small Agent action prediction accuracies in maze navigation. \acronym significantly outperforms baselines.}
    \label{tab:maze_result}
   \end{minipage}
    \end{minipage}
\end{figure*}

We compare \acronym{}s with a Boltzmann model in which agents select actions according to:
\begin{equation}
    Q_{\text{temp}}(a \mid s) = \xtemp \cdot \max_{\tau : \tau_0 = a} R(\tau)
\end{equation}
where $R(\tau)$ denotes the \emph{final} reward obtained along the complete trajectory $\tau$ (i.e.\ upon reaching some exit $R_i$). We also compare to simple baselines in which the agent performs truncated search up to a constant (not inferred) depth. We evaluate these models in two ways:

\paragraph{Predicting actions.}
In held-out states, we evaluate models' \textbf{exact-match} accuracy in predicting an agent's next action.
Results are shown in \cref{tab:maze_result}. Models that assume a constant depth perform worst. While Boltzmann models are better able to predict agents' next actions than these fixed-budget models, they are significantly outperformed by \acronym.
\vspace{-0.5em}
\paragraph{Predicting rewards.}
\vspace{-0.5em}
We also evaluate whether inferred prior distributions over $\beta$ recover the true values used to generate the data. Results for \acronym and the Boltzmann model are shown in \cref{fig:maze_temp}. It can be seen that \acronym almost perfectly recovers these parameters (suggesting that prediction errors in \cref{tab:maze_result} result entirely from errors in the inferred reward parameters $R_i$). Meanwhile, the Boltzmann model shows no significant differences in inferred $\xtemp$ across depth budgets, emphasizing the discrepancy between the two mdoels of suboptimality.

Together, these results show that \acronym is computationally tractable and capable of making accurate predictions and inferring meaningful parameters in simple search problems. In the remainder of the paper, we apply it to modeling real human behavior in more complex decision-making tasks.

\setlength{\textfloatsep}{5pt}

\setlength{\textfloatsep}{15pt}

\vspace{-0.5em}
\section{Pragmatic Language Understanding with Rational Speech Acts}
\label{sec:prag}

\begin{wraptable}{r}{0.3\textwidth}
\vspace{-1em}
\centering
\setlength{\tabcolsep}{2pt}
\resizebox{\linewidth}{!}{
\begin{tabular}[c]{r@{. \ } ccc r}
  \toprule
  \multicolumn{4}{c}{Context} & Utterance \\
  \midrule
  1&\colorContext{5866A7}{2DD2BC}{C23D5A}{purple}\\
  2&\colorContext{5866A7}{9953AC}{2DD2A6}{blue}\\
  3&\colorContext{3884C7}{02F9FD}{9E6461}{blue}\\
  \bottomrule
\end{tabular}
}
\caption{Example of the reference color (within the black box) and the two distractor colors, along with the utterance produced by a speaker from the colors in context task \citep{monroe2017colors}. Notice how the context affects the utterance, even as the reference color remains fixed. }
\label{tab:cic_examples}
\end{wraptable}

The next task we consider focuses on \textbf{pragmatic language understanding}---inferring speakers' communicative intents from their utterances. Humans readily produce and understand language in ways that deviates from its ``literal'' meaning. In \cref{tab:cic_examples}, for example, a color that would be described on its own by most speakers \emph{purple} is instead labeled \emph{blue} in some contexts.
A large body of work in cognitive science models this kind of context-based language understanding as the result of an iterative inference process \citep{frank2012predicting,franke2013game}: for example, in Row 2 of \cref{tab:cic_examples}, a speaker might choose to describe the highlighted color as \emph{blue} by reasoning that a na\"ive listener might resolve \emph{purple} to the second color in the row. A more sophisticated listener, in turn, can predict this speaker behavior, and successfully infer the intended meaning. But this kind of recursive reasoning about other agents can be computationally demanding, and requires sophisticated internal models of other language users. Experimental evidence suggests that is deployed selectively, and to different degrees by different language users \citep{franke2016reasoning}. We use L-IBMs to determine when, and to what extent, this kind of recursive reasoning is used during language production.

Our experiments focus on a \textbf{reference game} of the kind depicted in \cref{tab:cic_examples} \citep{monroe2017colors}. Reference games are a staple of research on pragmatic language use.
In a reference game, both a listener and speaker observe a set of candidate referents (e.g.\ colors). The speaker is privately given one of the colors as a target; they must then produce a natural language utterance for the listener. Finally, the listener selects a color patch, and both players win if they agreed on the target.

By fitting an \acronym to utterances and choices in human reference games, we investigate (1) whether we can infer whether humans are engaged in pragmatic reasoning from behavior alone, (2) whether there are differences between players in their ability to reason about their interlocutors, and (3) whether these differences actually predict communicative success (i.e.\ whether players with greater inference budgets are better at making themselves understood).

\subsection{Agent Model}
We build on the Rational Speech Acts (RSA) model of  \citet{frank2012predicting}. This model frames communication as one in which Bayesian listeners and speakers reason recursively about each others' beliefs in order to select utterances and actions. The starting point of RSA is a \textbf{literal listener}
$\xinference^{0}_{L}$ that maps utterances $u$ to actions according to their non-contextual meanings. (In \cref{tab:cic_examples}, 
a literal listener hearing the word \emph{purple} might choose randomly between the first two colors in the second row, as both would be reasonably described as purple out of context.)
The literal listener may be implemented by any model (e.g.\ a lookup table or a neural network; \citealp{andreas2016reasoning}) with parameters $\theta$.
Next, given a reference target $t$, a \textbf{pragmatic speaker}
$\xinference_{S}$ chooses an utterance in proportion to the probability that it will cause a literal listener to take the right action:
\begin{align}
 \xinference_{S}^1(u \mid t) \propto %
 p(\pi^0_L \textrm{ selects } t \textrm{ upon hearing } u) = \xinference^{0}_L(t \mid u)
  \label{eq:pragmatic_speaker}
\end{align}
(RSA speakers are standardly parameterized with an additional Boltzmann rationality parameter, which we will discuss momentarily.)
Finally, \textbf{pragmatic listeners} observe speaker utterances $u$, and reason about which reference targets were most likely to have produced those utterances:
\begin{align}
    \xinference_L^1(t \mid u) = p(\pi^1_S \textrm{ intends to signal } t \mid u) \propto \xinference^{1}_S(u \mid t) ~ p(t)
     \label{eq:pragmatic_listener}
\end{align}
Crucially, this process may be repeated, with speakers $\pi_S^i$ reasoning about ever-more-sophisticated speakers $\pi_L^{i-1}$, etc. %
But how many rounds of iteration actually explain human behavior? 
In the \algo framework, we may model this by embedding RSA inside an \acronym, with the budget $\beta$ parameterizing the number RSA iterations performed by each agent:
\begin{align}
    \pi_S(u \mid t; \theta, \eta) &= \sum_\beta \xdepthbudget(\beta \mid \eta) \pi_S(u \mid t; \theta, \beta) \\
    \pi_S(u \mid t; \theta, \beta) &= \pi_S^\beta(u \mid t)
\end{align}
(and analogously for $\pi_L$.)
\begin{proposition}
Rational Speech Acts (RSA) is an anytime inference algorithm. (Each inference state $f_\beta$ is $\pi_S^\beta$ or $\pi_L^\beta$. Each of these can be derived from the other in constant time via Eqs.\ref{eq:pragmatic_speaker}--\ref{eq:pragmatic_listener}.)
\end{proposition}

\subsection{Data}
For this task, we use the data collected by \cite{monroe2017colors}. Each game consists of roughly 50 rounds played between a human speaker and a human listener. In each round, the speaker observes a target color along with two distractors. The speaker produces an utterance and the listener has to click on one of the colors. The dataset consists of 46,994 rounds across 948 games. We create a 80/10/10 split across train, valid and test sets. \citeauthor{monroe2017colors} stratify the dataset into three difficulties (easy, medium and difficult) based on perceptual similarity between colors and distractors. 
Because each game is annotated with a unique identifier for both the speaker and the listener, we may further stratify the dataset according to \emph{player skill}: we compute the fraction of games won by each (speaker, listener) pair, then group these pairs into six buckets according to their win rate percentile relative to other players. This allows us to examine the relationship between inference budget and both task difficulty and communicative success.

\subsection{Models}

Following \cite{monroe2017colors}, we implement the literal listener $\pi_L^0$ using a transformer model that receives all three colors (represented as HSL vectors) and a natural language utterance as input, and predicts the index of the target color as output. We embed this listener model within the speaker--listener recursion defined by \cref{eq:pragmatic_listener}, then train it end-to-end (with budget parameters $\eta_i$) on the Colors in Context data using \cref{eq:objective}.

The constant of proportionality in \cref{eq:pragmatic_speaker} involves a sum over all natural language strings, which is cannot be computed efficiently. Here, also following \citep{monroe2017colors}, we perform a sampling-based approximation: we train a \emph{literal speaker} model to generate plausible utterances, then sum over a finite number of such samples to obtain a distribution over strings. See \citet{mcdowell2019learning} for more details. The literal speaker is parameterized identically to the literal listener, but outputs strings rather than color indices.

In experiments investigating the relationship between task difficulty and inference budget, we fit one $\eta_i$ \emph{per condition} (easy, medium, hard). In experiments investigating the relationship between communicative success and inference budget, we fit one $\eta_i$ \emph{per skill level} (between 1 and 6).

\subsection{Evaluation}

Standard implementations of RSA modifies \cref{eq:pragmatic_speaker} to include a Boltzmann parameter for speakers:
\begin{equation}
    \pi_S^i(u \mid t; \beta) \propto \exp \{\xtemp \log \pi_L^{i-1}(t \mid u) \}
    \label{eq:new_pragmatic_speaker}
\end{equation}
Like our $\xdepthbudget$, this parameter is intended to model possibly sub-optimal behavior on the part of speakers and listeners. We compare an \acronym to a model of this form. In particular, we fix the number of RSA iterations to one, use the same data as above to estimate literal listener parameters jointly with a prior distribution over $\xtemp$:
\begin{equation}
\pi_S^1(u \mid t; \theta, \eta) = \sum_\beta p_\text{temp}(\beta \mid \eta) \pi_S^1(u \mid t; \beta)
\end{equation}
where $\pi_S^1$ is defined as in \cref{eq:new_pragmatic_speaker}.

\cref{tab:cic_speaker} shows different models' ability to predict the target referent given human speaker utterances. Consistent with the findings of \citep{monroe2017colors}, 
because even literal models have access to all three referents, all model variants can achieve good task performance. When we look at inferred values for $\xdepthbudget$ and $\xtemp$, however, we begin to see significant differences between models. When stratifying games by \emph{difficulty}, we infer that the non-literal speaker is employed only for the hardest conditions. When stratifying games by \emph{player skill}, we infer that the weakest players can be modeled exclusively as literal speakers, while stronger players can be modeled as a mix of literal and pragmatic speakers. To the best of our knowledge, this is the first example of an RSA-type model being used to infer individual differences in pragmatic language use within a speaker population; we expect that these tools may be of independent interest to the cognitive science community. Additional experiments, predicting the object that the listener picked instead of the one the speaker is presented can be found in \cref{app:cic}.

\begin{figure*}[t]
\centering

\begin{minipage}[t]{0.45\textwidth}
\centering \footnotesize \textbf{Inferred} $\xtemp$
\end{minipage}%
\vspace{1em}
\begin{minipage}[t]{0.5\textwidth}
\centering \footnotesize \textbf{Inferred} $\xdepthbudget$ (L-IBM)
\end{minipage}
\vspace{1em}
\begin{minipage}[t]{0.2\textwidth}
  \centering
  \includegraphics[width=\linewidth]{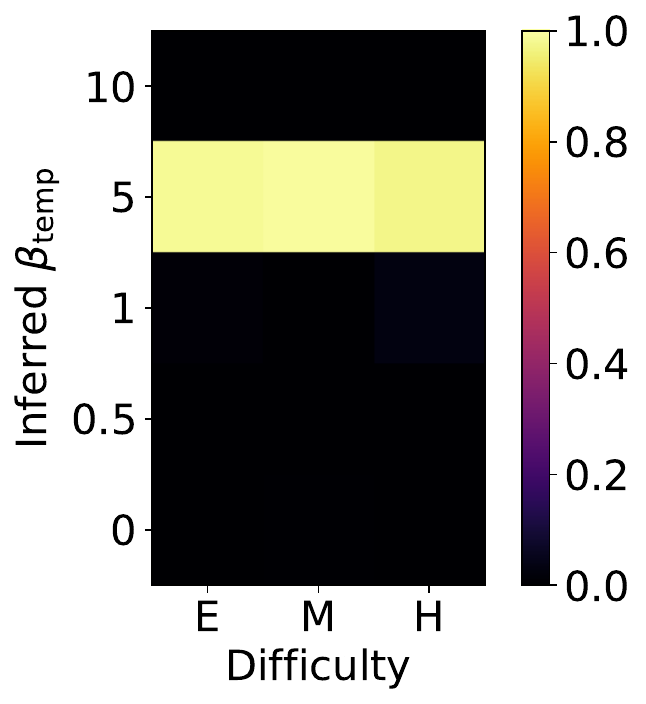}
  \subcaption{}
\end{minipage}%
\hfill 
\begin{minipage}[t]{0.25\textwidth}
  \centering
  \includegraphics[width=\linewidth]{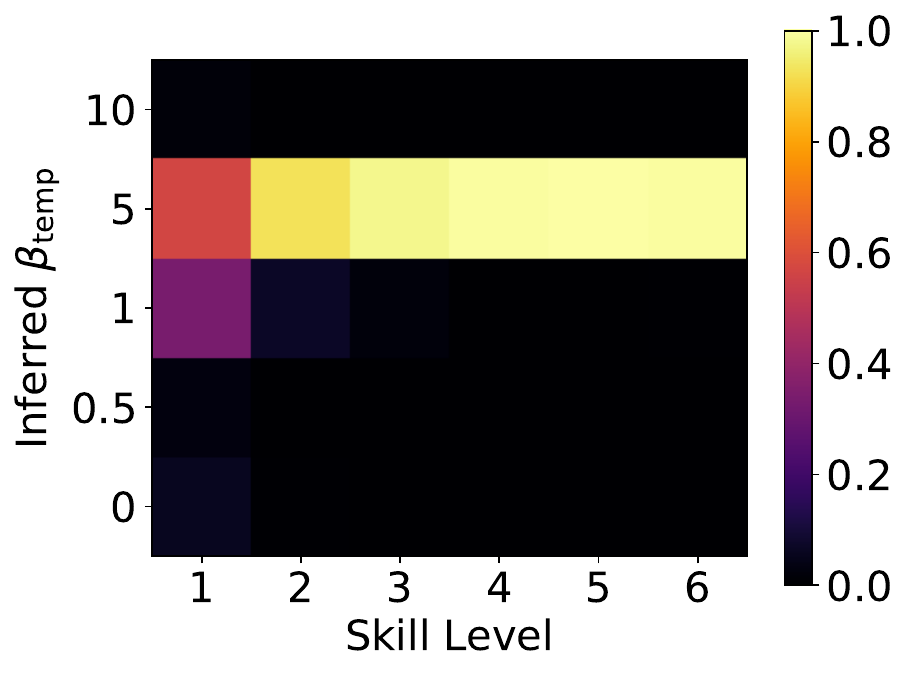}
\subcaption{}
\end{minipage}%
\hfill \vline \hfill
\begin{minipage}[t]{0.25\textwidth}
  \centering
  \includegraphics[width=\linewidth]{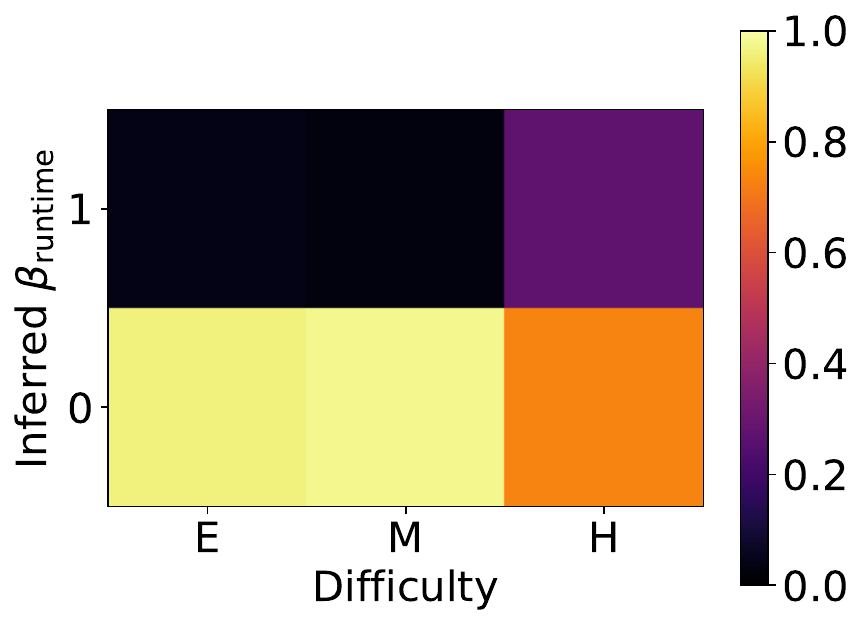}
    \subcaption{}
\end{minipage}%
\hfill
\begin{minipage}[t]{0.25\textwidth}
  \centering
  \includegraphics[width=\linewidth]{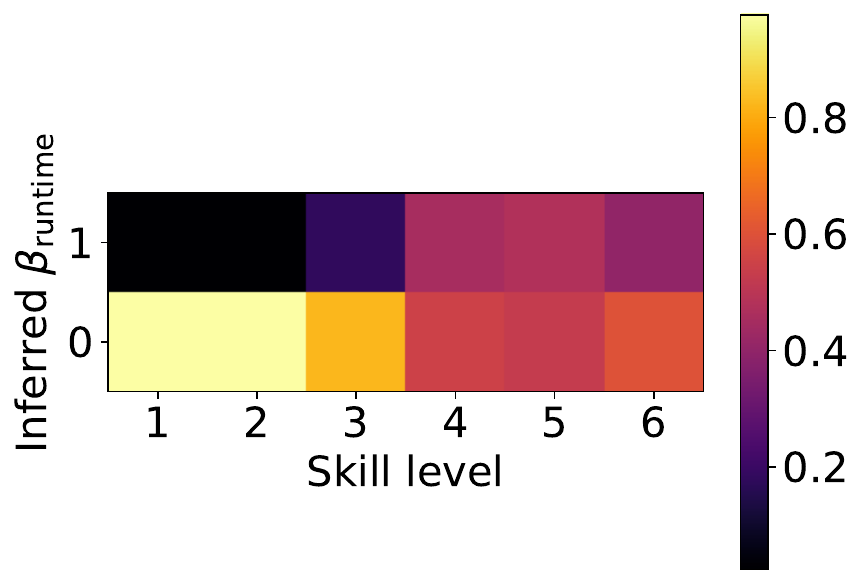}
\subcaption{}
\end{minipage}
\vspace{-1em}
\captionof{figure}{Inferred distributions over $\xbudget$ in RSA. X-axis indicates the difficulty level (\textbf{E}asy, \textbf{M}edium, \textbf{H}ard) or the player skill level (between 1 and 6, 6 being the most skilled players). The inferred $\xtemp$ across difficulty in a) and player skill in b) is not as meaningful as it is for $\xdepthbudget$ in c) and d). c) When separating games by difficulty, L-IBM infers that the non-literal speaker is employed only for the hardest condition. d) When separating games by player skill, we infer that the weakest players can be modeled exclusively as literal speakers, while stronger players can be modeled as a mix of literal and pragmatic speakers.}
\label{fig:cic_speaker}
\end{figure*}

\begin{figure*}[ht]
\centering
\scalebox{1.0}{
\centering
\begin{minipage}[t]{0.47\textwidth}
\vspace{-4.5em}
\centering
\resizebox{\linewidth}{!}{
\begin{tabular}{lcc}
\toprule
\textbf{Model} & \textbf{Type} & \textbf{Accuracy} \\
\midrule
$\xdepthbudget=0$  & -    & 83.3                   \\
$\xdepthbudget=1$   & -  & 83.0                     \\
\hline
\hline
Inferred $\xtemp$ &  player skill & 83.9 \\
Inferred $\xdepthbudget$ (\acronym) & player skill & \textbf{84.0} \\
\hline
Inferred $\xtemp$ & difficulty & 83.5 \\
Inferred $\xdepthbudget$ (\acronym) & difficulty & 82.7 \\
\bottomrule
\end{tabular}
}
\captionof{table}{Performance of different RSA models in predicting the speaker target. All models (including literal models and fixed-depth RSA models) achieve similar predictive performance—because even literal models have access to all three referents, all model variants can achieve good task performance. Note that $\xdepthbudget=0$ represents the base literal listener.}
\label{tab:cic_speaker}
\end{minipage}%
\hspace{0.6cm} %
\begin{minipage}[t]{0.47\textwidth}
\centering
\resizebox{\linewidth}{!}{
\begin{tabular}{lcc}
    \toprule
    \textbf{Model} & \textbf{Type} & \textbf{Accuracy} \\
    \midrule
    IL & - & 42.06 \\
    $\xdepthbudget=100$ & - & 43.64 \\
    \hline
    \hline
    Inferred $\xc$ & Active Elo & 43.77 \\
    Inferred $\xdepthbudget$ (\acronym) & Active Elo & \textbf{44.17} \\
    \hline
    Inferred $\xc$ & Opponent Elo & 43.84 \\
    Inferred $\xdepthbudget$ (\acronym) & Opponent Elo & \textbf{44.17} \\
    \hline
    Inferred $\xc$ & Time Control & 43.61 \\
    Inferred $\xdepthbudget$(\acronym) & Time Control & \bf 44.15 \\
    \bottomrule
\end{tabular}
}
\captionof{table}{Accuracy of predicting an agent's next action in chess. Models with MCTS outperform the depth-0 (imitation learning) baseline. Learning sub-population-specific $\xbudget$ enhances performance, with L-IBM-based learning of $\xdepthbudget$ consistently outperforming $\xc$ by a slight margin.}
\label{tab:chessresults}
\end{minipage}
}
\vspace{-1em}
\end{figure*}

\vspace{-0.5em}
\section{Playing Chess with Monte-Carlo Tree Search}
\vspace{-0.5em}
\label{sec:chess}

Finally, we turn from cooperative to adversarial decision-making tasks. 
We focus on chess, a popular two-player sequential game widely used as a benchmark for AI systems. 
Here, we are interested in modeling human chess play---specifically, observing data from a population of sub-optimal agents with a common reward function (winning the game) and attempting to infer those agents' computational constraints.
In human human play, there can be numerous sources of such constraints: a player paired against a strong opponent will likely to plan for longer than against a weaker opponent; some variants (like blitz chess) deliberately limit players' time-per-move (and, we might expect, the quality of their plans).
Given a dataset of human games played under different time constraints and player strengths, can we use \acronym{} to model variability in players' decisions across game states?

\subsection{Agent Model}

In this work, we model chess players as selecting actions using \textbf{Monte Carlo tree search} (MCTS).
Recent work \citep{jacob2022modeling} has shown that MCTS is a good model of strong human players. Here, following \citep{silver2017mastering,silver2016mastering, jacob2022modeling, grill2020monte}, we implement one of the most common modern forms of MCTS, which uses a value function $V$ predicting the expected total future reward 
and a policy prior $\xinference^0$ to guide exploration. 
At a high level, MCTS operates by incrementally growing a game tree starting at the root node, repeatedly picking some path to explore down the tree, performing a value function evaluation and then walking back up the tree updating all the value estimates based on that result. 
At each node, MCTS treats action selection as a multi-armed bandit problem. 
We use a standard exploration policy \citep{kocsis2006bandit}: during inference at each node of the search tree, we choose actions according to:
\begin{equation}
    \argmax_a \, Q_t(a \mid s) + \xc\xinference^0(a \mid s)\frac{\sqrt{\sum_b N(s,b)}}{N(s,a)+1}
\end{equation}
where $Q_t(s,a)$ is the estimated expected future reward for $i$ from playing action $a$ in state $s$ at iteration $t$, the visit count $N(s,a)$ is the number of times $a$ has been explored from $s$, $\xinference^0(a \mid s)$ is an ``anchor'' policy, and $\xbudget_{\text{puct}}$ is a tunable parameter trading off exploration versus exploitation. 
After expanding $\xdepthbudget$ nodes of this tree, an agent's final action is sampled from a distribution:
\begin{align}
    \label{eq:mcts}
    \xinference(a \mid s; \xdepthbudget) = \xc \frac{\sqrt{\xdepthbudget}}{N(s,a)+1} \frac{\xinference^0(a|s)}{\gamma - Q_{\xdepthbudget}(a \mid s)}
\end{align}
where $\gamma$ is chosen such that $\xinference$ forms a proper probability distribution. 
\begin{proposition}
Monte-Carlo tree search (MCTS) is an anytime inference algorithm. 
(Let each inference state $f_\beta$ be the tree of nodes and visitation counts after $\beta$ evaluations. This tree is refined by evaluating \cref{eq:mcts} once.)
\end{proposition}
With $\pi(a \mid s; \xdepthbudget)$ as defined above, we may instantiate an \acronym for MCTS:
\begin{align}
  \xourmodel^{\text{runtime}}(t | u;\eta, \theta) &= \sum_{\xdepthbudget} \budgetdist(\xdepthbudget\mid\eta_i) \cdot \xinference(a ; s, \xdepthbudget)
\end{align}
\vspace{-0.5em}
\subsection{Data}
\vspace{-0.5em}
We use similar data to previous models of human chess play by \citet{mcilroy2020aligning,jacob2022modeling,mcilroy2022learning}. 
Our experiments use two different datasets.
First, a dataset $D_\text{large}$ containing roughly 6 million moves; second, a dataset $D_\text{small}$ containing roughly 75,000 moves. $D_\text{small}$ includes metadata describing players' Elo ratings (a measure of strength) and game formats (the amount of time players had to select moves). See \cref{app:chess_data} for details.

\subsection{Modeling details}
We train the base initial policy $\xinference_0$ and a value model $\tilde v_0$ as two different output heads of a deep neural network using imitation learning on the large dataset split $D_{\text{large}}$. Our architecture is a 4-block residual network similar to those used in prior work \citep{mcilroy2020aligning,jacob2022modeling,mcilroy2022learning}.
Unlike previous sections, we do not learn the value functions jointly with $p_\text{budget}$.
Instead, we first learn a single value function from $D_{\text{large}}$, then
fit $\budgetdist(\xc \mid\eta_i)$ and $\budgetdist(\xdepthbudget \mid\eta_i)$.
We investigate three ways of stratifying players into sub-populations: 
player Elo (a proxy for player skill), and opponent Elo and time control (both proxies for task difficulty). As in \cref{sec:prag}, we estimate a separate $\eta_i$ for each group within each stratified dataset.

\subsection{Evaluation}
\label{subsec:chesseval}
Unlike in the two domains studied above, there is already an established literature on modeling sub-optimal behavior via MCTS outside the Boltzmann framework. The most successful current approach models individual differences in play \citep{jacob2022modeling} by fitting $\beta_\text{puct}$. We thus compare to a baseline in which $\eta_i$ parameterizes a distribution over values of $\xc$ rather than tree expansions.

\begin{figure*}[t]
\centering
\vspace{-1em}
\begin{minipage}[b]{\textwidth}
  \centering \small \textbf{Inferred} $\xc$
\end{minipage}
\begin{minipage}[b]{0.33\textwidth}
  \centering
  \includegraphics[width=\linewidth]{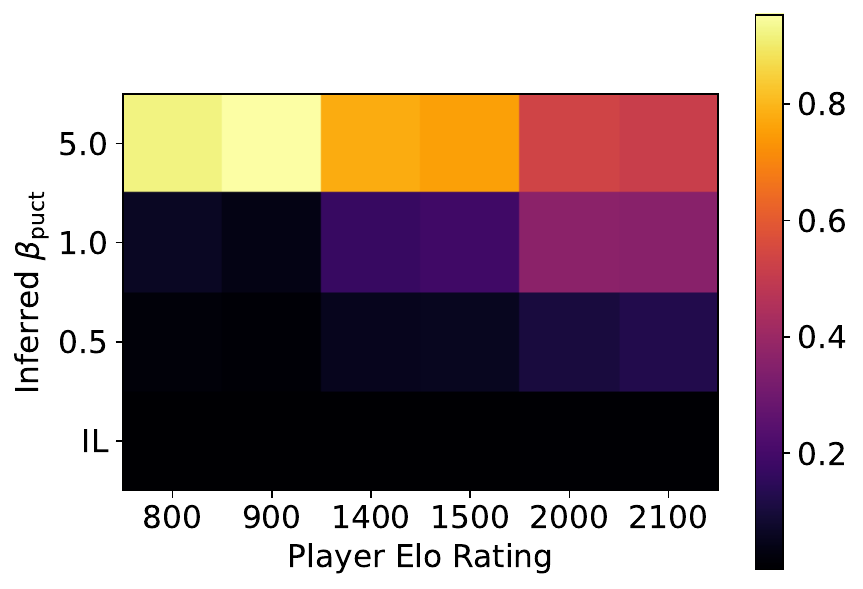}
\end{minipage}%
\begin{minipage}[b]{0.33\textwidth}
  \centering
  \includegraphics[width=\linewidth]{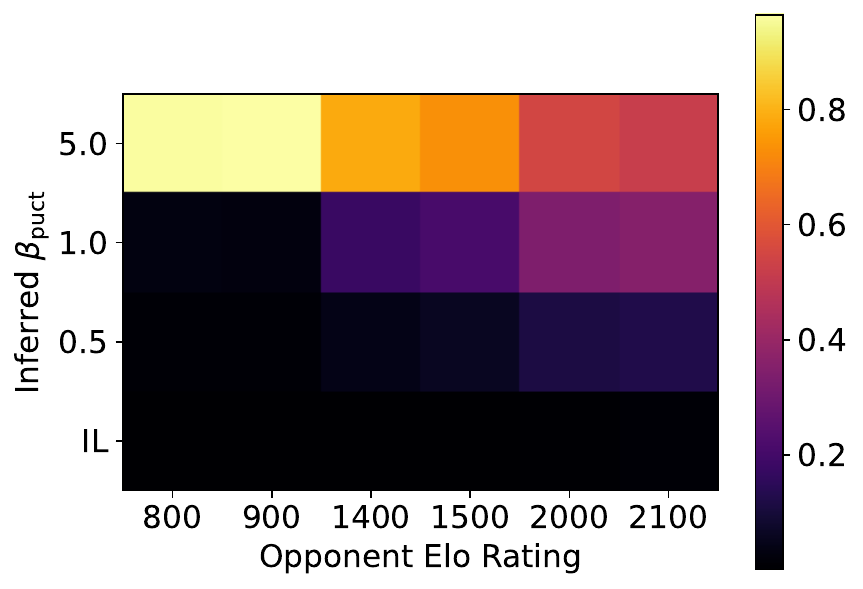}
\end{minipage}%
\begin{minipage}[b]{0.3\textwidth}
  \centering
  \includegraphics[width=\linewidth]{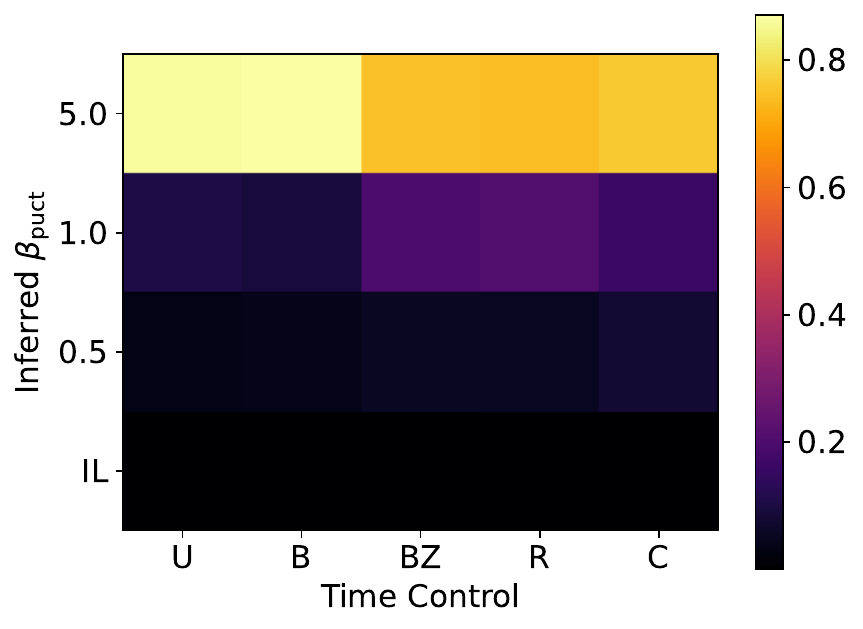}
\end{minipage}%
\vspace{1em}
\begin{minipage}[b]{\textwidth}
  \centering \small \textbf{Inferred} $\xdepthbudget$ (L-IBM)
\end{minipage}
\begin{minipage}[b]{0.33\textwidth}
  \centering
  \includegraphics[width=\linewidth]{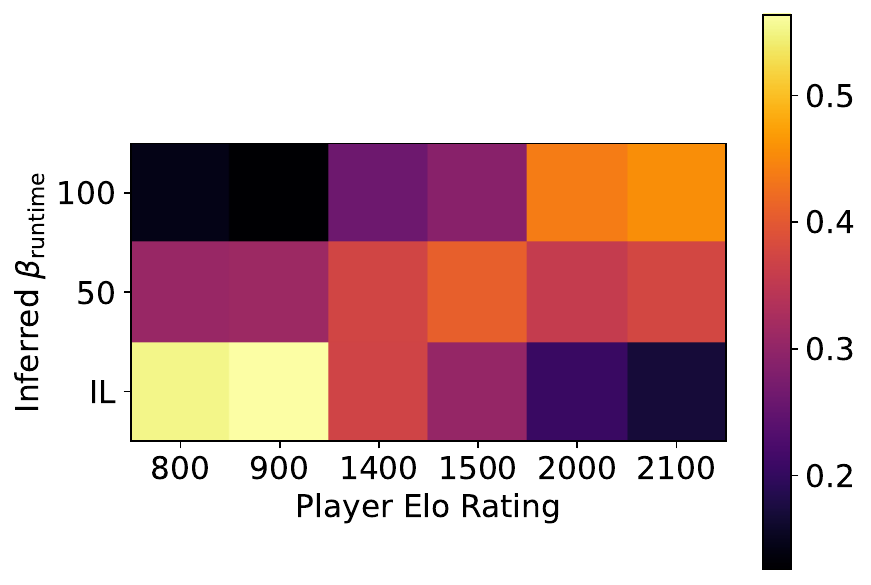}
\end{minipage}%
\begin{minipage}[b]{0.33\textwidth}
  \centering
  \includegraphics[width=\linewidth]{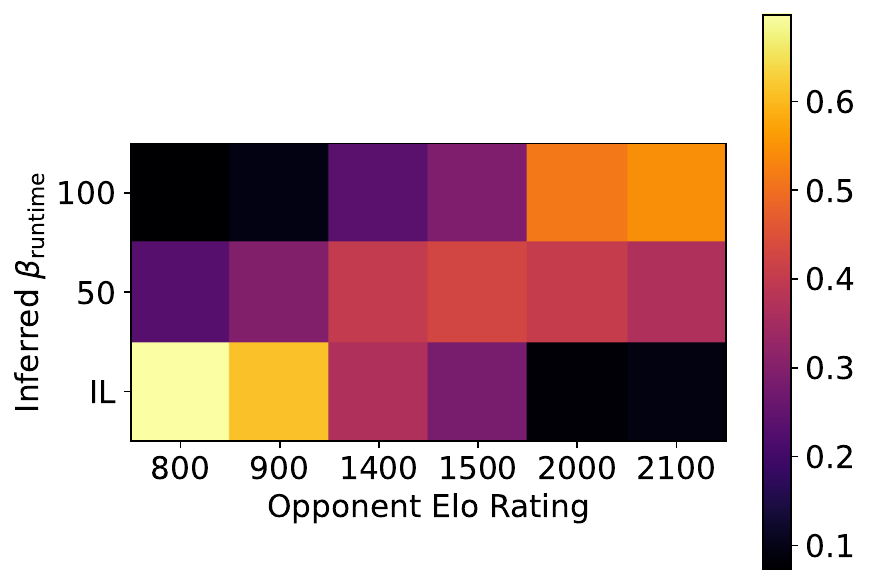}
\end{minipage}%
\begin{minipage}[b]{0.3\textwidth}
  \centering
  \includegraphics[width=\linewidth]{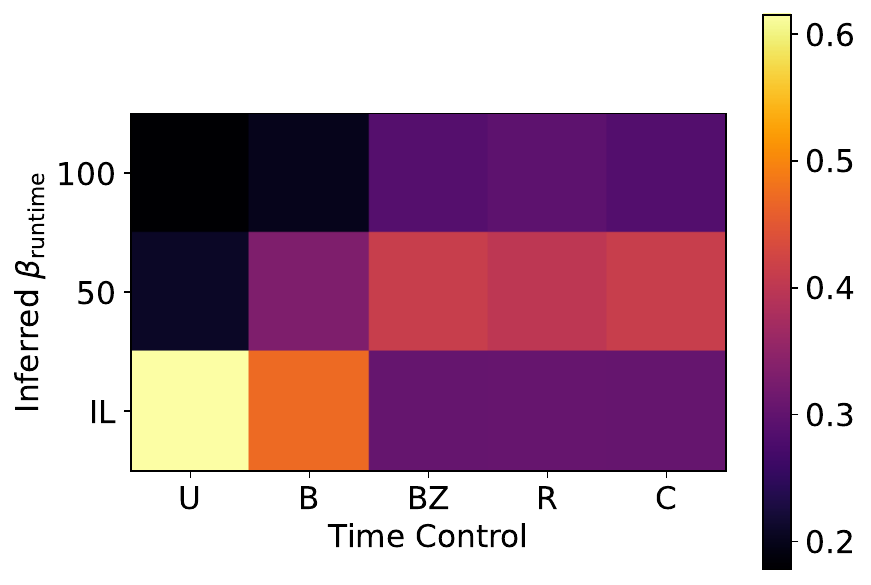}
\end{minipage}

\caption{Inferred distributions over $\xbudget$ in Chess using MCTS. X-axis indicates the player Elo rating, opponent elo rating buckets and time control: Ultra Bullet (U), Bullet (B), Blitz (BZ), Rapid (R) and Classical (C). The top row depicts the distributions for $\xc$ and the bottom row depicts the distributions for $\xdepthbudget$. When the player's or opponent's strength increases, $\xdepthbudget$ infers greater runtime. This pattern also holds true as the time control extends. $\xc$ displays a similar pattern, as the agents or opponents get stronger, or as the time control extends, $\xc$ suggests lower values, placing greater reliance on the search Q-values.}
\label{fig:chess}
\end{figure*}

Accuracy (in terms of top-one predictions and negative log-likelihood) is reported in \cref{tab:chessresults}.
As in past work, we find that models that with explicit search outperform imitation-learning baseline. Learning sub-population specific $\xbudget$ improves the performance even further, with \acronym-based learning of $\xdepthbudget$ consistently outperforming $\xc$ by a small margin. 

Inferred budget parameters are shown in \cref{fig:chess}.
Here, we observe that as the player strength or the opponent strength increases as measured by the Elo ratings, $\xdepthbudget$ infers higher runtime. We also observe the same as the time control increases: $\xdepthbudget$ infers higher runtime as the duration of each move of the game increases. 
$\xc$ shows a weaker, but similar trend:
as the agents or opponents get stronger, or as the time control increases, $\xc$ infers lower values of $\xc$, indicating that players are deviating from the prior and are relying more on the search Q-values.

\section{Conclusion}

We have described \algo{}s, a family of approaches for modeling agents acting to achieve unknown goals subject to unknown constraints on their inferential capabilities. Instead of assuming either global optimality of decision-making or uniform suboptimality, our approach explicitly infers the runtime that agents devote to approximate inference.
This paradigm is applicable to all anytime inference algorithms. 
In three domains---maze navigation, pragmatic language understanding, and playing chess---we demonstrated that it can outperform classical models of bounded rationality while imputing meaningful measures of human skill and task difficulty.

\section*{Acknowledgements}

This work was supported by the National Science Foundation under grants IIS-2238240 and IIS-2212310. Thanks to Jennifer Hu for helpful discussions about modeling individual differences in Rational Speech Act models. We also thank David Wu and Noam Brown for helpful discussions about Monte-Carlo tree search.


\begin{thebibliography}{34}
\providecommand{\natexlab}[1]{#1}
\providecommand{\url}[1]{\texttt{#1}}
\expandafter\ifx\csname urlstyle\endcsname\relax
  \providecommand{\doi}[1]{doi: #1}\else
  \providecommand{\doi}{doi: \begingroup \urlstyle{rm}\Url}\fi

\bibitem[Andreas \& Klein(2016)Andreas and Klein]{andreas2016reasoning}
Jacob Andreas and Dan Klein.
\newblock Reasoning about pragmatics with neural listeners and speakers.
\newblock In \emph{Proceedings of the Conference on Empirical Methods in
  Natural Language Processing}, pp.\  1173--1182, 2016.

\bibitem[Beliaev et~al.(2022)Beliaev, Shih, Ermon, Sadigh, and
  Pedarsani]{beliaev2022imitation}
Mark Beliaev, Andy Shih, Stefano Ermon, Dorsa Sadigh, and Ramtin Pedarsani.
\newblock Imitation learning by estimating expertise of demonstrators.
\newblock In \emph{Proceedings of the International Conference on Machine
  Learning}, pp.\  1732--1748. PMLR, 2022.

\bibitem[Boddy \& Dean(1989)Boddy and Dean]{boddy1989solving}
Mark Boddy and Thomas~L Dean.
\newblock \emph{Solving time-dependent planning problems}.
\newblock Brown University, Department of Computer Science, 1989.

\bibitem[Callaway et~al.(2022)Callaway, van Opheusden, Gul, Das, Krueger,
  Griffiths, and Lieder]{callaway2022rational}
Frederick Callaway, Bas van Opheusden, Sayan Gul, Priyam Das, Paul~M Krueger,
  Thomas~L Griffiths, and Falk Lieder.
\newblock Rational use of cognitive resources in human planning.
\newblock \emph{Nature Human Behaviour}, 6\penalty0 (8):\penalty0 1112--1125,
  2022.

\bibitem[Camerer et~al.(2004)Camerer, Ho, and Chong]{camerer2004cognitive}
Colin~F Camerer, Teck-Hua Ho, and Juin-Kuan Chong.
\newblock A cognitive hierarchy model of games.
\newblock \emph{The Quarterly Journal of Economics}, 119\penalty0 (3):\penalty0
  861--898, 2004.

\bibitem[Dean \& Boddy(1988)Dean and Boddy]{anytime1988}
Thomas~L Dean and Mark~S Boddy.
\newblock An analysis of time-dependent planning.
\newblock In \emph{Proceedings of the Annual Meeting of the Association for the
  Advancement of Artificial Intelligence}, volume~88, pp.\  49--54, 1988.

\bibitem[Devlin et~al.(2019)Devlin, Chang, Lee, and Toutanova]{devlin2018bert}
Jacob Devlin, Ming-Wei Chang, Kenton Lee, and Kristina Toutanova.
\newblock {BERT}: {P}re-training of deep bidirectional transformers for
  language understanding.
\newblock In \emph{Proceedings of the Conference of the North American Chapter
  of the Association for Computational Linguistics: Human Language
  Technologies}, pp.\  4171--4186, 2019.

\bibitem[Frank \& Goodman(2012)Frank and Goodman]{frank2012predicting}
Michael~C Frank and Noah~D Goodman.
\newblock Predicting pragmatic reasoning in language games.
\newblock \emph{Science}, 336\penalty0 (6084):\penalty0 998--998, 2012.

\bibitem[Franke(2013)]{franke2013game}
Michael Franke.
\newblock Game theoretic pragmatics.
\newblock \emph{Philosophy Compass}, 8\penalty0 (3):\penalty0 269--284, 2013.

\bibitem[Franke \& Degen(2016)Franke and Degen]{franke2016reasoning}
Michael Franke and Judith Degen.
\newblock Reasoning in reference games: Individual-vs. population-level
  probabilistic modeling.
\newblock \emph{PloS one}, 11\penalty0 (5):\penalty0 e0154854, 2016.

\bibitem[Griffiths et~al.(2019)Griffiths, Callaway, Chang, Grant, Krueger, and
  Lieder]{griffiths2019doing}
Thomas~L Griffiths, Frederick Callaway, Michael~B Chang, Erin Grant, Paul~M
  Krueger, and Falk Lieder.
\newblock Doing more with less: {Meta}-reasoning and meta-learning in humans
  and machines.
\newblock \emph{Current Opinion in Behavioral Sciences}, 29:\penalty0 24--30,
  2019.

\bibitem[Grill et~al.(2020)Grill, Altch{\'e}, Tang, Hubert, Valko, Antonoglou,
  and Munos]{grill2020monte}
Jean-Bastien Grill, Florent Altch{\'e}, Yunhao Tang, Thomas Hubert, Michal
  Valko, Ioannis Antonoglou, and R{\'e}mi Munos.
\newblock {Monte-Carlo} tree search as regularized policy optimization.
\newblock In \emph{In Proceedings of the International Conference on Machine
  Learning}, pp.\  3769--3778. PMLR, 2020.

\bibitem[Haarnoja et~al.(2017)Haarnoja, Tang, Abbeel, and
  Levine]{haarnoja2017reinforcement}
Tuomas Haarnoja, Haoran Tang, Pieter Abbeel, and Sergey Levine.
\newblock Reinforcement learning with deep energy-based policies.
\newblock In \emph{Proceedings of the International Conference on Machine
  Learning}, pp.\  1352--1361. PMLR, 2017.

\bibitem[Harris et~al.(2020)Harris, Millman, van~der Walt, Gommers, Virtanen,
  Cournapeau, Wieser, Taylor, Berg, Smith, Kern, Picus, Hoyer, van Kerkwijk,
  Brett, Haldane, del R{\'{i}}o, Wiebe, Peterson, G{\'{e}}rard-Marchant,
  Sheppard, Reddy, Weckesser, Abbasi, Gohlke, and Oliphant]{harris2020array}
Charles~R. Harris, K.~Jarrod Millman, St{\'{e}}fan~J. van~der Walt, Ralf
  Gommers, Pauli Virtanen, David Cournapeau, Eric Wieser, Julian Taylor,
  Sebastian Berg, Nathaniel~J. Smith, Robert Kern, Matti Picus, Stephan Hoyer,
  Marten~H. van Kerkwijk, Matthew Brett, Allan Haldane, Jaime~Fern{\'{a}}ndez
  del R{\'{i}}o, Mark Wiebe, Pearu Peterson, Pierre G{\'{e}}rard-Marchant,
  Kevin Sheppard, Tyler Reddy, Warren Weckesser, Hameer Abbasi, Christoph
  Gohlke, and Travis~E. Oliphant.
\newblock Array programming with {NumPy}.
\newblock \emph{Nature}, 585\penalty0 (7825):\penalty0 357--362, September
  2020.
\newblock URL \url{https://doi.org/10.1038/s41586-020-2649-2}.

\bibitem[Huys et~al.(2012)Huys, Eshel, O'Nions, Sheridan, Dayan, and
  Roiser]{huys2012bonsai}
Quentin~JM Huys, Neir Eshel, Elizabeth O'Nions, Luke Sheridan, Peter Dayan, and
  Jonathan~P Roiser.
\newblock Bonsai trees in your head: {H}ow the {P}avlovian system sculpts
  goal-directed choices by pruning decision trees.
\newblock \emph{PLoS Computational Biology}, 8\penalty0 (3):\penalty0 e1002410,
  2012.

\bibitem[Huys et~al.(2015)Huys, Lally, Faulkner, Eshel, Seifritz, Gershman,
  Dayan, and Roiser]{huys2015interplay}
Quentin~JM Huys, N{\'\i}all Lally, Paul Faulkner, Neir Eshel, Erich Seifritz,
  Samuel~J Gershman, Peter Dayan, and Jonathan~P Roiser.
\newblock Interplay of approximate planning strategies.
\newblock \emph{Proceedings of the National Academy of Sciences}, 112\penalty0
  (10):\penalty0 3098--3103, 2015.

\bibitem[Jacob et~al.(2022)Jacob, Wu, Farina, Lerer, Hu, Bakhtin, Andreas, and
  Brown]{jacob2022modeling}
Athul~Paul Jacob, David~J Wu, Gabriele Farina, Adam Lerer, Hengyuan Hu, Anton
  Bakhtin, Jacob Andreas, and Noam Brown.
\newblock Modeling strong and human-like gameplay with {KL-R}egularized search.
\newblock In \emph{Proceedings of the International Conference on Machine
  Learning}, pp.\  9695--9728. PMLR, 2022.

\bibitem[Jeon et~al.(2020)Jeon, Milli, and Dragan]{jeon2020reward}
Hong~Jun Jeon, Smitha Milli, and Anca Dragan.
\newblock Reward-rational (implicit) choice: A unifying formalism for reward
  learning.
\newblock In \emph{Advances in Neural Information Processing Systems}, pp.\
  4415--4426, 2020.

\bibitem[Kingma \& Ba(2015)Kingma and Ba]{kingma2014adam}
Diederik Kingma and Jimmy Ba.
\newblock Adam: A method for stochastic optimization.
\newblock In \emph{Proceedings of the International Conference on Learning
  Representations}, 2015.

\bibitem[Kocsis \& Szepesv{\'a}ri(2006)Kocsis and
  Szepesv{\'a}ri]{kocsis2006bandit}
Levente Kocsis and Csaba Szepesv{\'a}ri.
\newblock Bandit based {Monte-Carlo} planning.
\newblock In \emph{Proceedings of European Conference on Machine Learning},
  pp.\  282--293. Springer, 2006.

\bibitem[Liang et~al.(2017)Liang, Liaw, Nishihara, Moritz, Fox, Gonzalez,
  Goldberg, and Stoica]{liang2017ray}
Eric Liang, Richard Liaw, Robert Nishihara, Philipp Moritz, Roy Fox, Joseph
  Gonzalez, Ken Goldberg, and Ion Stoica.
\newblock Ray rllib: A composable and scalable reinforcement learning library.
\newblock \emph{arXiv preprint arXiv:1712.09381}, 85, 2017.

\bibitem[Luce(1959)]{luce2012individual}
R~Duncan Luce.
\newblock Individual choice behavior.
\newblock 1959.

\bibitem[McDowell \& Goodman(2019)McDowell and Goodman]{mcdowell2019learning}
Bill McDowell and Noah Goodman.
\newblock Learning from omission.
\newblock In \emph{Proceedings of the Annual Meeting of the Association for
  Computational Linguistics}, pp.\  619--628, 2019.

\bibitem[McIlroy-Young et~al.(2020)McIlroy-Young, Sen, Kleinberg, and
  Anderson]{mcilroy2020aligning}
Reid McIlroy-Young, Siddhartha Sen, Jon Kleinberg, and Ashton Anderson.
\newblock Aligning superhuman ai with human behavior: Chess as a model system.
\newblock In \emph{Proceedings of the ACM SIGKDD International Conference on
  Knowledge Discovery \& Data Mining}, pp.\  1677--1687, 2020.

\bibitem[McIlroy-Young et~al.(2022)McIlroy-Young, Wang, Sen, Kleinberg, and
  Anderson]{mcilroy2022learning}
Reid McIlroy-Young, Russell Wang, Siddhartha Sen, Jon Kleinberg, and Ashton
  Anderson.
\newblock Learning models of individual behavior in chess.
\newblock In \emph{Proceedings of the 28th ACM SIGKDD Conference on Knowledge
  Discovery and Data Mining}, pp.\  1253--1263, 2022.

\bibitem[Meltzoff(1995)]{meltzoff1995understanding}
Andrew~N Meltzoff.
\newblock Understanding the intentions of others: Re-enactment of intended acts
  by 18-month-old children.
\newblock \emph{Developmental Psychology}, 31\penalty0 (5):\penalty0 838, 1995.

\bibitem[Monroe et~al.(2017)Monroe, Hawkins, Goodman, and
  Potts]{monroe2017colors}
Will Monroe, Robert~XD Hawkins, Noah~D Goodman, and Christopher Potts.
\newblock Colors in context: A pragmatic neural model for grounded language
  understanding.
\newblock \emph{Transactions of the Association for Computational Linguistics},
  5:\penalty0 325--338, 2017.

\bibitem[Paszke et~al.(2019)Paszke, Gross, Massa, Lerer, Bradbury, Chanan,
  Killeen, Lin, Gimelshein, Antiga, et~al.]{paszke2019pytorch}
Adam Paszke, Sam Gross, Francisco Massa, Adam Lerer, James Bradbury, Gregory
  Chanan, Trevor Killeen, Zeming Lin, Natalia Gimelshein, Luca Antiga, et~al.
\newblock {PyTorch}: An imperative style, high-performance deep learning
  library.
\newblock \emph{Advances in Neural Information Processing Systems}, 32, 2019.

\bibitem[Raffel et~al.(2020)Raffel, Shazeer, Roberts, Lee, Narang, Matena,
  Zhou, Li, and Liu]{raffel2020exploring}
Colin Raffel, Noam Shazeer, Adam Roberts, Katherine Lee, Sharan Narang, Michael
  Matena, Yanqi Zhou, Wei Li, and Peter~J Liu.
\newblock Exploring the limits of transfer learning with a unified text-to-text
  transformer.
\newblock \emph{The Journal of Machine Learning Research}, 21\penalty0
  (1):\penalty0 5485--5551, 2020.

\bibitem[Russell \& Wefald(1991)Russell and Wefald]{russell1991principles}
Stuart Russell and Eric Wefald.
\newblock Principles of metareasoning.
\newblock \emph{Artificial Intelligence}, 49\penalty0 (1-3):\penalty0 361--395,
  1991.

\bibitem[Silver et~al.(2016)Silver, Huang, Maddison, Guez, Sifre, Van
  Den~Driessche, Schrittwieser, Antonoglou, Panneershelvam, Lanctot,
  et~al.]{silver2016mastering}
David Silver, Aja Huang, Chris~J Maddison, Arthur Guez, Laurent Sifre, George
  Van Den~Driessche, Julian Schrittwieser, Ioannis Antonoglou, Veda
  Panneershelvam, Marc Lanctot, et~al.
\newblock Mastering the game of {Go} with deep neural networks and tree search.
\newblock \emph{Nature}, 529\penalty0 (7587):\penalty0 484--489, 2016.

\bibitem[Silver et~al.(2018)Silver, Hubert, Schrittwieser, Antonoglou, Lai,
  Guez, Lanctot, Sifre, Kumaran, Graepel, et~al.]{silver2017mastering}
David Silver, Thomas Hubert, Julian Schrittwieser, Ioannis Antonoglou, Matthew
  Lai, Arthur Guez, Marc Lanctot, Laurent Sifre, Dharshan Kumaran, Thore
  Graepel, et~al.
\newblock A general reinforcement learning algorithm that masters chess, shogi,
  and {Go} through self-play.
\newblock \emph{Science}, 362\penalty0 (6419):\penalty0 1140--1144, 2018.

\bibitem[van Opheusden et~al.(2023)van Opheusden, Kuperwajs, Galbiati, Bnaya,
  Li, and Ma]{nature2023expertise}
Bas van Opheusden, Ionatan Kuperwajs, Gianni Galbiati, Zahy Bnaya, Yunqi Li,
  and Wei~Ji Ma.
\newblock Expertise increases planning depth in human gameplay.
\newblock \emph{Nature}, pp.\  1--6, 2023.

\bibitem[Wolf et~al.(2020)Wolf, Debut, Sanh, Chaumond, Delangue, Moi, Cistac,
  Rault, Louf, Funtowicz, et~al.]{wolf2019huggingface}
Thomas Wolf, Lysandre Debut, Victor Sanh, Julien Chaumond, Clement Delangue,
  Anthony Moi, Pierric Cistac, Tim Rault, R{\'e}mi Louf, Morgan Funtowicz,
  et~al.
\newblock Transformers: State-of-the-art natural language processing.
\newblock In \emph{Proceedings of the Conference on Empirical Methods in
  Natural Language Processing: System Demonstrations}, pp.\  38--45, 2020.

\end{thebibliography}
\newpage
\appendix
\section{Training hyperparameters}
\label{sec:app}
We will detail the training hyperparameter details in this section.
\subsection{Maze}
\label{app:maze_hyper}
All models in \cref{sec:maze} were trained using the Adam optimizer \citep{kingma2014adam}, where the learning rates were sweeped across the following values $[1.0, 0.5, 1e-1, 0.05, 1e-2, 5e-3, 1e-3, 5e-4, 1e-4, 5e-5]$ for 50 epochs. The values presented in \cref{tab:maze_result} were picked from the model with the best validation accuracy across the learning rates. We implemented truncated BFS using Pytorch \citep{paszke2019pytorch} and Numpy \citep{harris2020array} and we also used the \href{https://github.com/john-science/mazelib}{mazelib library} to generate the data.

\subsection{Colors in context}
The models trained in \cref{sec:prag} are based on the transformer architecture and trained from scratch. The speaker model was trained based on the T5 model \citep{raffel2020exploring} with the following hyperparameters described in \cref{tab:t5config}. The speaker was trained with a batch size of 64 using the Adam optimizer with learning rate $1e-4$ for 25 epochs. 

\begin{table*}[h]
\begin{minipage}{.4\linewidth}
  \centering
  \resizebox{\linewidth}{!}{
    \begin{tabular}{lc}
     \toprule
      \textbf{Parameter} & \textbf{Value} \\
      \midrule
      Number of Layers & 4 \\
      Number of Heads & 4 \\
      Model Dimension & 32 \\
      Key-Value Dimension & 16 \\
      Feedforward Dimension & 32 \\
      \bottomrule
    \end{tabular}
  }
  \caption{Hyperparameter configuration of the speaker model based on T5 \citep{raffel2020exploring}.}
  \label{tab:t5config}
\end{minipage}%
\hspace{2cm}
\begin{minipage}{.4\linewidth}
  \centering
  \resizebox{\linewidth}{!}{
    \begin{tabular}{lc}
      \toprule
      \textbf{Parameter} & \textbf{Value} \\
      \midrule
      Hidden Size & 64 \\
      Number of Hidden Layers & 4 \\
      Number of Attention Heads & 4 \\
      Intermediate Hidden Size & 256 \\
      \bottomrule
    \end{tabular}
  }
  \caption{Hyperparameter configuration of the listener model based on BERT.}
  \label{tab:bertconfig}
\end{minipage}
\end{table*}

All the listener models were based on the BERT \citep{devlin2018bert} model with the configuration described in \cref{tab:bertconfig}. The listener models were trained using Adam and the learning rates were sweeped across the following values $[1e-3, 5e-4, 1e-4, 5e-5]$ for upto 50 epochs. The values presented in \cref{tab:cic_listener} and \cref{tab:cic_speaker} were picked from the model with the best validation accuracy across the learning rates. We trained the models using Pytorch \citep{paszke2019pytorch} and Huggingface \citep{wolf2019huggingface} libraries. We specifically implemented RSA using Pytorch.

\subsection{Chess}
The value and policy network used in \cref{sec:chess} are based on an architecture that is a 4-block residual network similar to those used in prior work \cite{mcilroy2020aligning,jacob2022modeling,mcilroy2022learning}. The policy and value network was trained using Adam with a learning rate of 0.001, a batch size of 4096 and for upto 30 epochs.  The epoch used in the rest of the section was picked based on the validation accuracy.

In the second set of fine-tuning experiments, for every set of conditioning type, a simple feedforward network was trained using Adam with a batch size of 512. The models in \cref{sec:chess} were picked by selecting the learning rates between $1e-3, 5e-4, 1e-4, 5e-5$ with the best validation accuracy.

For chess, the base policy and value functions were trained using Ray library \citep{liang2017ray} and Pytorch. MCTS was specifically implemented using Numpy. We also used the \href{https://pettingzoo.farama.org/}{pettingzoo library} for simulating moves.

\section{Chess Data}
\label{app:chess_data}

$D_\text{large}$ consists of 5,974,872 moves in the training split, 60,968 in the validation split and 60,969 moves in the test set. These data points were randomly sampled from the January, 2019 database release of a chess website (lichess). $D_\text{small}$ consists of 50,000 moves in the training split, 12,041 moves in the validation split and 12,040 moves in the test split. These data points were randomly sampled from the February, 2019 lichess database release but filtering such that only those players with Elo ratings in the following buckets were considered: [800-1000], [1400-1600] and [2000-2200]. 

The dataset contains 5 different types of time control.
In increasing duration, they are \textbf{Ultra Bullet}, \textbf{Bullet}, \textbf{Blitz}, \textbf{Rapid} and \textbf{Classical} (see \cref{table:time_control}).

\begin{table*}[h]
\centering
\begin{tabular}{lc}
\toprule
\textbf{Time control} & \textbf{Estimated Duration (seconds)} \\
\midrule
UltraBullet & $< 29$ \\
Bullet & $< 179$ \\
Blitz & $< 479$ \\
Rapid & $< 1499$ \\
Classical & $\geq 1500$ \\
\bottomrule
\end{tabular}
\caption{Estimated game durations across different time controls.}
\label{table:time_control}
\end{table*}

The time controls used in our work have estimated durations that are defined in \cref{table:time_control}:

\section{Additional Experiments: Colors in context}
\label{app:cic}
\begin{figure*}[h]
\centering

\begin{minipage}[b]{0.2\textwidth}
  \centering
  \includegraphics[width=\linewidth]{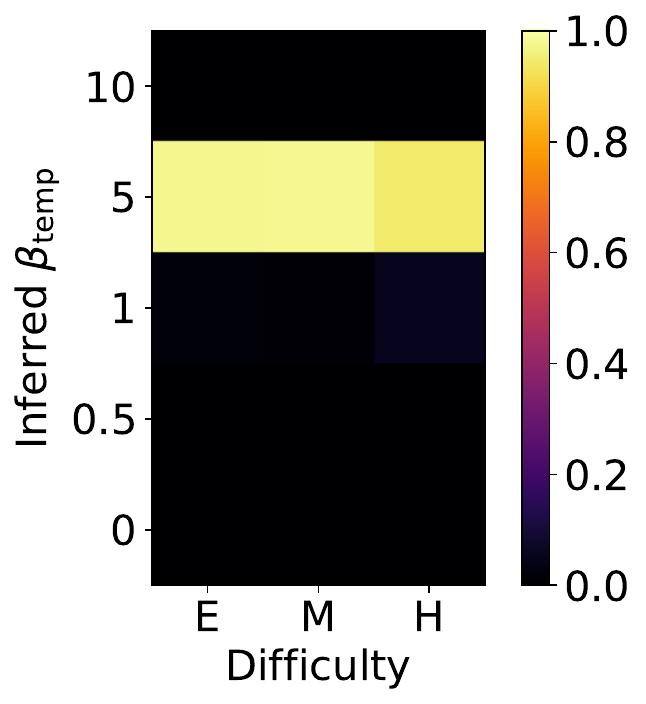}
  \subcaption{}
\end{minipage}%
\hfill
\begin{minipage}[b]{0.25\textwidth}
  \centering
  \includegraphics[width=\linewidth]{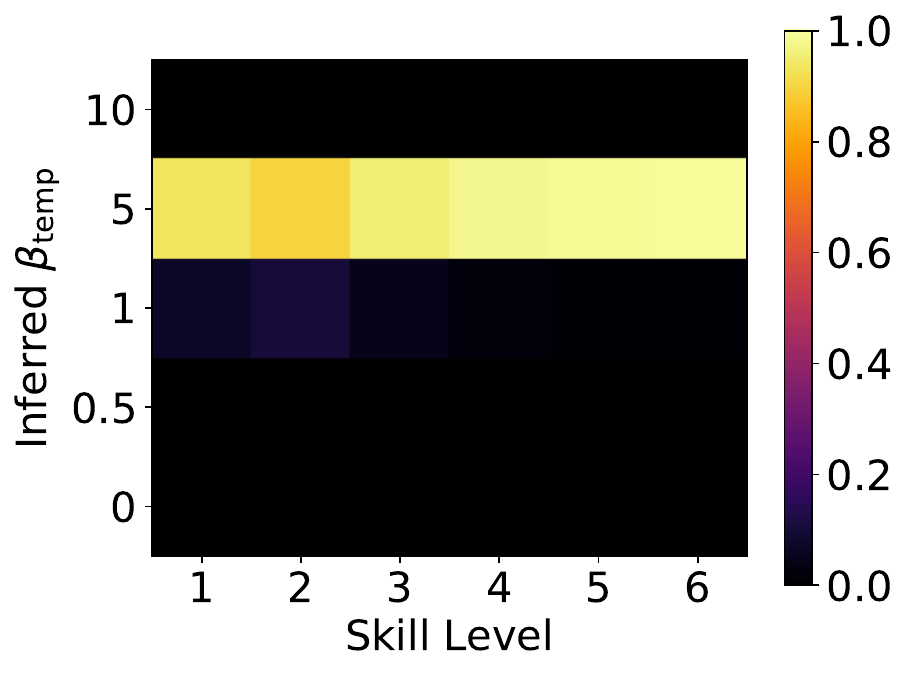}
\subcaption{}
\end{minipage}%
\hfill
\begin{minipage}[b]{0.25\textwidth}
  \centering
  \includegraphics[width=\linewidth]{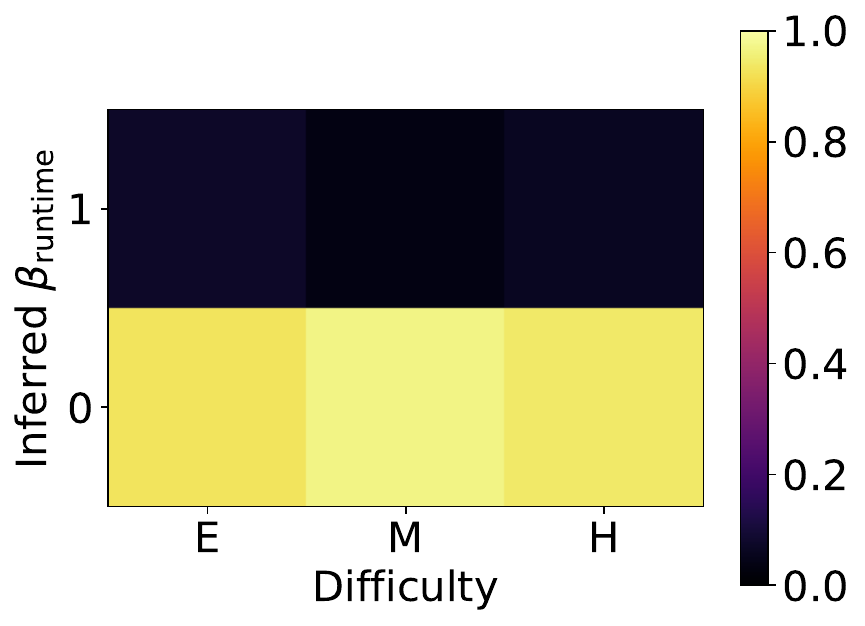}
    \subcaption{}
\end{minipage}%
\hfill
\begin{minipage}[b]{0.25\textwidth}
  \centering
  \includegraphics[width=\linewidth]{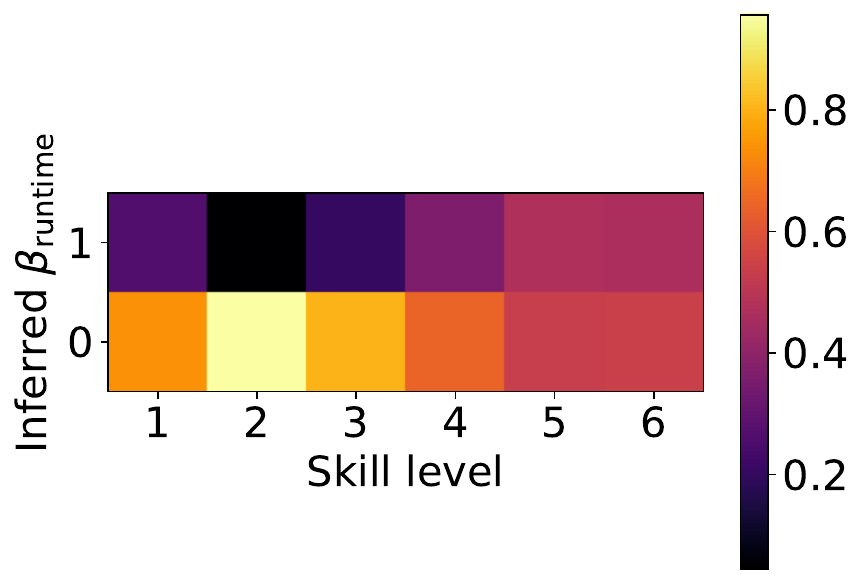}
\subcaption{}
\end{minipage}

\captionof{figure}{Inferred distributions over $\xbudget$ in RSA, with the listener target. X-axis indicates the difficulty level (\textbf{E}asy, \textbf{M}edium, \textbf{H}ard) or the player skill level (1 - 6, 6 being the most skilled players). The inferred $\xtemp$ across difficulty in a) and player skill in b) is not as meaningful as it is for $\xdepthbudget$ in d). When separating games by player skill, we infer that the weakest players can be modelled with a smaller mix towards pragmatic speakers compared to stronger players}
\label{fig:cic_listener}
\end{figure*}

In this section, we include additional experiments for the pragmatics domain where we train the models to predict the object that the listener picks. We present the results of a similar set of experiments as in \cref{sec:prag} in \cref{tab:cic_listener} and \cref{fig:cic_listener}. We specifically note that the inference based approaches outperform the baselines in this setting. 

\begin{table*}[h]
\centering
\begin{tabular}{lcc}
\toprule
\textbf{Model} & \textbf{Type} & \textbf{Accuracy} \\
\midrule
$\xdepthbudget=0$ (Literal listener)   & -    & 80.4                   \\
$\xdepthbudget=1$   & -  & 81.8                     \\
\hline
Inferred $\xtemp$ &  player skill & 82.3 \\
Inferred $\xdepthbudget$ (\acronym) & player skill & \textbf{83.1} \\
\hline
Inferred $\xtemp$ & difficulty & 82.7 \\
Inferred $\xdepthbudget$ (\acronym) & difficulty & 82.1 \\
\bottomrule
\end{tabular}
\captionof{table}{Performance of different RSA models in predicting the speaker target. The $\xbudget$ based models outperform the baseline models: literal models and fixed-depth RSA models.}
\label{tab:cic_listener}
\end{table*}

\section{Additional Discussion}
\subsection{Relationship between $\beta_\mathrm{puct}$ and ELO rating in chess}
$\beta_\mathrm{puct}$ (as used in popular strength-modeling approaches like \cite{silver2016mastering}) doesn’t simply control exploration vs. exploitation. Instead, it biases exploration towards an initial policy prior $\pi^0$. As $\beta_\mathrm{puct}$ tends to infinity, it is identical to playing $\pi^0$. When $\beta_\mathrm{puct}$ is set to 0, it is equivalent to greedily picking the search Q-values. In \cref{fig:chess} and as it relates to ELO rating, we notice that stronger players start deviating more from their base policy $\pi^0$ to instead depend more on their MCTS search Q-values. Therefore indicating that stronger players rely more on search compared to weaker players.

\end{document}